\pdfoutput=1

\documentclass[11pt]{article}

\usepackage[preprint]{acl}

\usepackage{times}
\usepackage{latexsym}

\usepackage[T1]{fontenc}

\usepackage[utf8]{inputenc}

\usepackage{microtype}

\usepackage{inconsolata}

\usepackage{graphicx}

\usepackage{mathtools}
\usepackage{hyperref}
\usepackage{caption}
\usepackage{subcaption}
\usepackage{multirow}
\usepackage{tabularx, array, booktabs}

\title{Show or Tell? Modeling the evolution of request-making in Human-LLM conversations}

\author{Shengqi Zhu, 
  Jeffrey M. Rzeszotarski, 
  David Mimno \\
  Department of Information Science \\
  Cornell University \\
  \texttt{\{sz595, mimno\}@cornell.edu}, \texttt{jeff.rzeszotarski@gmail.com} \\}

\begin{document}
\maketitle
\begin{abstract}
Designing user-centered LLM systems requires understanding how people use them, but patterns of user behavior are often masked by the variability of queries.
In this work, we introduce a new framework to describe request-making that segments user input into request content, roles assigned, query-specific context, and the remaining \textit{task-independent} expressions.
We apply the workflow to create and analyze a dataset of 211k real-world queries based on WildChat.
Compared with similar human-human setups, we find significant differences in the language for request-making in the human-LLM scenario.
Further, we introduce a novel and essential perspective of \textit{diachronic analyses} with user expressions, which reveals fundamental and habitual user-LLM interaction patterns beyond individual task completion.
We find that query patterns evolve from early ones emphasizing sole requests to combining more context later on, and individual users explore expression patterns but tend to converge with more experience.
From there, we propose to understand communal trends of expressions underlying distinct tasks and discuss the preliminary findings.
Finally, we discuss the key implications for user studies, computational pragmatics, and LLM alignment.
\end{abstract}

\section{Introduction}

The versatile conversation format of chat-based LLM interfaces has created an unprecedented interaction paradigm by enabling open-ended user inputs~\cite{zhu2025data,10.1145/3613905.3650786}.
However, little attention has been given to the \textit{language} aspects of human-LLM interaction, much less the formation and evolution of users' language patterns over longer sessions: can we study \textit{how} people ask, independent of \textit{what} they are asking for?
More often, NLP studies focus on the semantics and type of tasks conveyed~\cite{tamkin2024clio,cheng2025realm,handa2025economic}, while the text modality itself is simply assumed as the default, easy way.
However, the natural language format is far more than a convenient form for tasks.
How users organize and contextualize their requests directly reflects an LLM's user-perceived affordances, expectations, and social roles.
Language does not just ``help to show how LLMs are used''; it \textit{is} how LLMs are used.
Specifically, users' habitual expressions generalized across tasks, a fundamental linguistic feature of user-LLM interaction, remain largely understudied.
Recent work observes real-world user-side nuances, such as users adapting their behaviors and expectations~\cite{choi-etal-2024-llm,schroeder2024large}, or changes in utterances upon model updates~\cite{10.1145/3613905.3651100}.
However, there is still yet to be a formal and generalizable framework to accommodate the linguistic analysis of input patterns beyond individual case studies.

In this work, we analyze how user inquiries are formed and explore the patterns in interactions over relative and absolute time.
First, we perform a new segmentation task to tackle the challenge of users freely embedding \textit{requests} in their \textit{expressions}, together with other parts with substantial presence like \textit{contexts} and assigned \textit{roles}.
Via an extendable, semi-automatic LLM annotation workflow, we present a dataset on top of WildChat~\cite{zhaowildchat} with 211,414 parsed user utterances consisting of Request Content, Context, Roles, and Expressions (ReCCRE). The dataset features the clean separation of the parts specific to a request from the generic natural language expressions used to deliver the request.

Using the dataset, we make several key observations and conclusions.
We show that the LLM chat modality is fundamentally different from natural request-making conversations, by comparing with the Stanford Politeness Datasets~\cite{danescu-niculescu-mizil-etal-2013-computational}, a primary resource in computational pragmatics.
We identify key repeating patterns in queries, which range on an axis between \textit{request-centric} and \textit{context-infused}.
More importantly, we introduce how the perspective of expressions enables \textit{diachronic} user modeling, with use records as a time-lapse data source for understanding user lifecycles and the community. We find clear traces that, as a user gains familiarity with the system, their expressions change less, and they migrate from simple chunks of requests to more context combinations. 
Further, we explore full-scale community analysis and discuss patterns over time supported by the data resource, showing that key trends are visible even with simple, non-parametric but indicative metrics like lexical diversity.
We conclude with the tangible implications and future directions regarding understanding users, pragmatics, and LLM alignment.
Our data, code, and prompts will be open-sourced.

\section{Related Work}

Interpreting Real-World Human-LLM Conversations has been a rising topic thanks to new data resources~\cite{zhaowildchat,zhenglmsys}. However, the major body of work has focused on the detection and categorization of what tasks users use LLMs for~\cite{zhang2025one,cheng2025realm,mireshghallahtrust} and their impacts~\cite{handa2025economic,tamkin2024clio,kirk2024prism}, as well as specific features such as values~\cite{huang2025values} and jailbreaking attempts~\cite{jin2025jailbreakhunter}. Beyond NLP, the formatting of prompts as well as the broader user experience with LLM input interfaces has also been core topics in Human-Computer Interaction~\cite{10.1145/3544548.3581388,10.1145/3706598.3714319,10.1145/3613905.3650786,zhang2025navigating}.

Some existing work shares the focus on individual attributes relevant to our discussions. \citet{lee2025realtalk} looks into evolving dialog patterns across time but focuses on inter-human and inter-LLM data synthesis instead of existing user-LLM documentations. For human-LLM, \citet{huang-etal-2024-characterizing} considers the ``conversational tones'' shared or (mis)aligned between humans and LLMs, and \citet{zheng-etal-2024-helpful} probes LLM performance with different role assignments. However, our work is fundamentally different: Prior work usually targets what a human-LLM conversation \textit{should} look like, a dominant thread related to fine-tuning and deployment~\cite{mott2024thing,mishra2022please,ivey2024real}; Our work focuses on the mining of existing data and new analysis paradigms. More importantly, we present a new systematic view that is not covered by the studies on single attributes.

At a high level, our work is most related to \citet{mysore2025prototypical} and \citet{kolawole2025parallelprompt}, which also seek to extract and describe latent patterns from massive user inputs. However, both works have distinct goals from this work: the former focuses on assisted writing and marks behavioral ``PATHs'' as a dialog proceeds, and the latter targets task taxonomies with semantic similarities.

\begin{figure}[t!]
    \centering
    \includegraphics[width=0.9\linewidth]{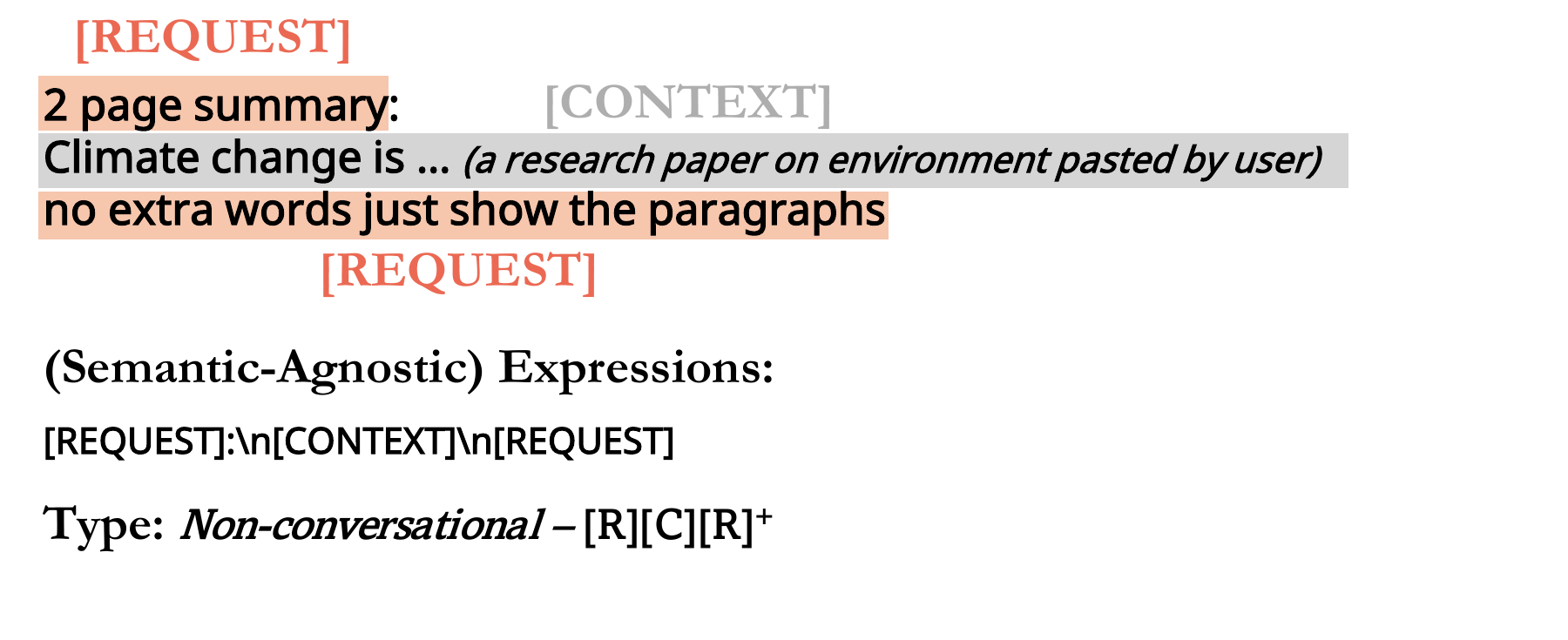}
    \includegraphics[width=0.9\linewidth]{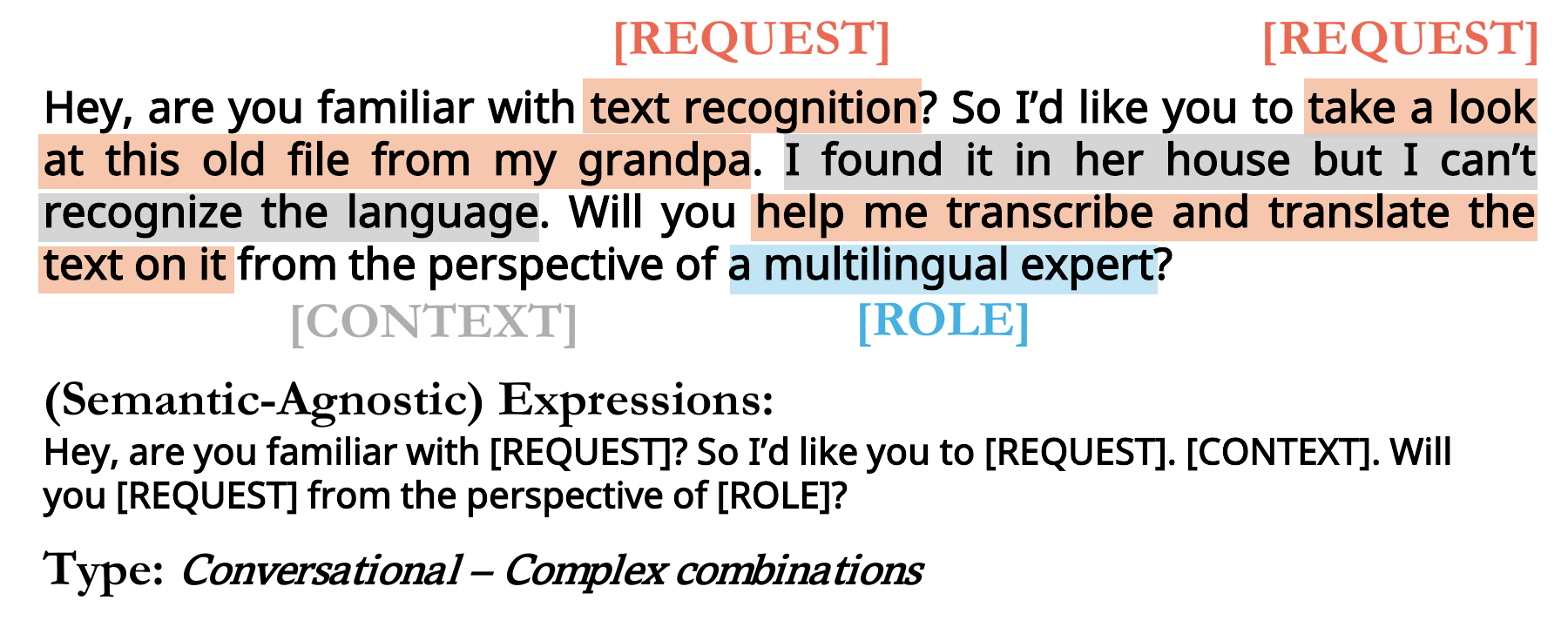}
    \caption{Two examples of user input annotated with request content, context, and roles. The precise definitions of the components are discussed in \S\ref{subsec:annotation_task_setup}, and the expression types are elaborated in Table~\ref{tab:expression_types}.}
    \label{fig:example}
\end{figure}

\section{Annotating User Request-Making}

We start by constructing the data infrastructure for modeling user request-making behaviors by segmenting requests.
Our goal is a generalizable annotation scheme that enables systematic quantitative and linguistic analysis of user requests, and allows adaptation to unseen data and to natural language conversations for comparisons.

\subsection{Task Setup}
\label{subsec:annotation_task_setup}

\paragraph{Source Data} 
We collect 317,373 initial user inputs (i.e., first turns) from the WildChat dataset~\cite{zhaowildchat} as the base corpus. 

\paragraph{Annotating User Queries}
As the first step, we distinguish between the making of an effortful request and the direct retrieval of answers (e.g., ``who is the 3rd president of the U.S.''). While the latter represents another common type of usage, the engagement level is low; it is less likely to involve a conversational scenario, and the dynamics are remarkably different from request-making.
In practice, the annotator first reads the input text thoroughly and determines whether it involves a request or a direct question, or ``both'' or ``neither''.

\begin{table*}[t!]
\centering
\resizebox{0.96\textwidth}{!}{
\begin{tabular}{|cl|l|l|}
\hline
\multicolumn{2}{|c|}{\textbf{Type}} & \multicolumn{1}{c|}{\textbf{Description}} & \multicolumn{1}{c|}{\textbf{Examples}} \\ \hline
\multicolumn{1}{|c|}{\multirow{8}{*}[-4.5em]{Non-conversational}} & $[R]$ & A single \textit{Request} component. & \begin{tabular}[c]{@{}l@{}}$[R]$\\ Give me $[R]$.\end{tabular} \\ \cline{2-4} 
\multicolumn{1}{|c|}{} & $[R]*n$ & \begin{tabular}[c]{@{}l@{}}Multiple \textit{Request}s concatenated in simple ways,\\ without conversational expressions.\end{tabular} & \begin{tabular}[c]{@{}l@{}}$[R]$ and $[R]$.\\ $[R]$. $[R]$. Also $[R]$.\end{tabular} \\ \cline{2-4} 
\multicolumn{1}{|c|}{} & $[R][C]$ & \begin{tabular}[c]{@{}l@{}}One \textit{Request} followed by One \textit{Context} component,\\ without conversational expressions.\end{tabular} & \begin{tabular}[c]{@{}l@{}}$[R]$: $[C]$\\ $[R]$ such as $[C]$.\end{tabular} \\ \cline{2-4} 
\multicolumn{1}{|c|}{} & $[C][R]$ & \begin{tabular}[c]{@{}l@{}}One \textit{Context} followed by One \textit{Request} component,\\ without conversational expressions.\end{tabular} & \begin{tabular}[c]{@{}l@{}}$[C]$. Now, $[R]$.\\ $[C]$\textbackslash{}n\textbackslash{}n$[R]$\end{tabular} \\ \cline{2-4} 
\multicolumn{1}{|c|}{} & $[R][C][R]^{+}$ & \begin{tabular}[c]{@{}l@{}}A pair of \textit{Request} and \textit{Context},\\ followed by additional \textit{Request} components.\end{tabular} & \begin{tabular}[c]{@{}l@{}}$[R]$, $[C]$. $[R]$ and $[R]$.\\ $[R]$: $[C]$. $[R]$. $[R]$. $[R]$.\end{tabular} \\ \cline{2-4} 
\multicolumn{1}{|c|}{} & $[R][C][C]^{+}$ & \begin{tabular}[c]{@{}l@{}}A pair of \textit{Request} and \textit{Context},\\ followed by additional \textit{Context} components.\end{tabular} & \begin{tabular}[c]{@{}l@{}}$[R]$ based of $[C]$: $[C]$.\\ $[R]$. $[C]$. $[C]$. $[C]$.\end{tabular} \\ \cline{2-4} 
\multicolumn{1}{|c|}{} & $[C]^{+}$ & \begin{tabular}[c]{@{}l@{}}Concatenation of \textit{Context} components only.\\ Usually seen in early requests to complete writings.\end{tabular} & \begin{tabular}[c]{@{}l@{}}$[C]$.\\ $[C]$ $[C]$\end{tabular} \\ \cline{2-4} 
\multicolumn{1}{|c|}{} & Other $[R]/[C]/[role]$ compositions & \begin{tabular}[c]{@{}l@{}}Other more complicated series of $[R]$, $[C]$, and $[role]$,\\ with simple, non-conversational expressions.\end{tabular} & \begin{tabular}[c]{@{}l@{}}$[R], [C]$. Then, $[R], [C]$. Finally, $[R]$, $[C]$.\\ $[C]$. $[C]$. Given $[C]$, $[R]$.\end{tabular} \\ \hline
\multicolumn{1}{|c|}{\multirow{3}{*}[-1.25em]{Conversational}} & Single $[R]$ & One single \textit{Request} in a conversational expression. & \begin{tabular}[c]{@{}l@{}}Can you help me to $[R]$?\\ Hi I wanna $[R]$.\end{tabular} \\ \cline{2-4} 
\multicolumn{1}{|l|}{} & Simple $[R]/[C]/[role]$ combinations & \begin{tabular}[c]{@{}l@{}}Simple combinations of \textit{Request}, \textit{Context}, and \textit{Role}\\ using conversational expressions.\end{tabular} & \begin{tabular}[c]{@{}l@{}}You are $[role]$. Now, $[R]$.\\ I'm working on $[C]$ and I'd like you to $[R]$.\end{tabular} \\ \cline{2-4} 
\multicolumn{1}{|l|}{} & Complex compositions & \begin{tabular}[c]{@{}l@{}}Other more complicated series of $[R]$, $[C]$, and $[role]$,\\ with full, conversational expressions.\end{tabular} & \begin{tabular}[c]{@{}l@{}}How can I $[R]$? $[R]$. Please be sure to $[R]$!\\ Act as $[role]$ and $[R]$. You will $[R]$: $[C]$.\end{tabular} \\ \hline
\end{tabular}
}
\caption{Taxonomy of user expressions, with 8 non-conversational types and 3 conversational types.}
\label{tab:expression_types}
\end{table*}

Next, for the request-making cases, we outline the case-specific core semantics relevant to the request.
This separates the framing templates used to deliver the request (expressions) across different dialogs.
Specifically, we introduce the following annotations of three elements of requests plus the user expressions, on top of the full user input text:

\begin{itemize}
    \item \textbf{Request Content ($[R]$)}: A core span of text that specifies what task(s) exactly the user wants the LLM to perform, or what goal(s) the user wants to achieve;
    \item \textbf{Context ($[C]$)}: A detailed span of context information that does not directly constitute the request, but provides support for neighboring requests. This includes the chunks of ``target text'' that the LLMs are requested to process (e.g., the pasted article for the request ``2 page summary'' in Fig.~\ref{fig:example}).
    \item \textbf{Role ($[role]$)}: Any roles that the LLM is asked to take on to achieve the requests.
    \item \textbf{Expression}: The remaining text after extracting the above components, which represent the generic language templates used to embed and deliver the requests.
\end{itemize}

Each word in a request is assigned to precisely one of the four categories, and the labels thus form a non-overlapping full division of the user input. This forms the basis of the corpus annotation, and we discuss next the implementation in full scale.

\subsection{Automating the annotation pipeline}
To adapt the annotation to chat logs at the million-request scale, we seek to balance between automation and reliability (consistency). 
We implement a review-and-revise pipeline that simultaneously generates semi-supervised annotations and extends the available data for fine-tuning with consistent standards learned from human annotation.

The pipeline involves three contributors: a human annotator; an interim SotA LLM specialized in text understanding ($L$); and a smaller local LLM as full-scale annotator ($l$).
We bootstrap from a small number of hand-labelled data, use $L$ as pseudo-reference for fine-tuning $l$, and eventually automate annotations with the fine-tuned local $l$.

First, the human annotator manually labeled a small, random batch of ``root'' data. This small collection of references is consistent and we denote this \textit{gold} dataset as $D_0$. Both LLM annotators are then provided with a detailed prompt of annotation rules and fine-tuned based on $D_0$. Next, both $L$ and $l$ are evaluated on a separate, randomly sampled set $D^{raw}_1$. We then convert this raw subset into a \textit{silver} set $D_1$ as follows:
\begin{itemize}
    \item If $L$ and $l$ \textit{agree} on the instance (the total difference in labels sufficiently low), the annotation of $L$ is accepted and added to $D_1$.
    \item Otherwise, if $L$ and $l$ \textit{disagree}, the human annotator reviews and selects an output, and if both are incorrect, the instance is manually labeled. The reviewed/relabelled version is added to $D_1$.
\end{itemize}
$D_1$ as a silver set is then merged with $D_0$ to form an expanded annotated dataset for fine-tuning. Both $L$ and $l$ are then fine-tuned and evaluated on another new batch $D_2^{raw}$, and the process is repeated so on and so forth.
Finally, after fine-tuning the models with incremental, semi-supervised data, we migrate the model prediction process from the black-box, high-cost $L$ to our local model $l$ to perform full-scale automated annotation.

We use \texttt{gpt-4o-2024-08-06} as the intermediate $L$, and a flagship 10B-level open-source LLM at the time of the work, \texttt{Qwen2.5-14B-Instruct}, as the full-scale annotator $l$. The review-and-revise loop was repeated 3 times on a total of 762 instances, and a template-based verification showed that ill-formatted responses are $<0.5\%$.

\section{The ReCCRE dataset}

\subsection{Basic information}
After obtaining the valid annotations, we set up a spam filter to remove consecutive dialogs that are overly similar or created within a very short period, and also remove the instances that do not involve request-making.
This results in a collection of 211,414 user request-making inputs from 18,964 users.
The dataset covers the time window from April 2023 to May 2024, and spans 6 versions of the \texttt{gpt-3.5-turbo} and \texttt{gpt-4} models.\footnote{See \citet{zhaowildchat} for the detailed documentation.}

\paragraph{Long-term Users}
To track the change of use patterns over time, we focus on the core users with sufficient experience and time to play with and adapt to the system.
In practice, we seek users (1) whose use records (time between first and last dialog created) span more than 14 days, and (2) have started at least 10 dialogs.
This yields a subset of 2,092 users with 59,175 request-making user inputs in total.
We refer to this key group as \textit{long-term} or \textit{stable} users.
Our analyses will primarily focus on this group, as it more reliably reflects use patterns and provides the legitimacy for diachronic observations.
We present a full-scale case study comparing long-term users with all users and the full data in \S\ref{subsec:community_mtld} under the Lexical Richness framework.

\begin{figure*}[t]
    \centering
    \includegraphics[width=0.88\linewidth]{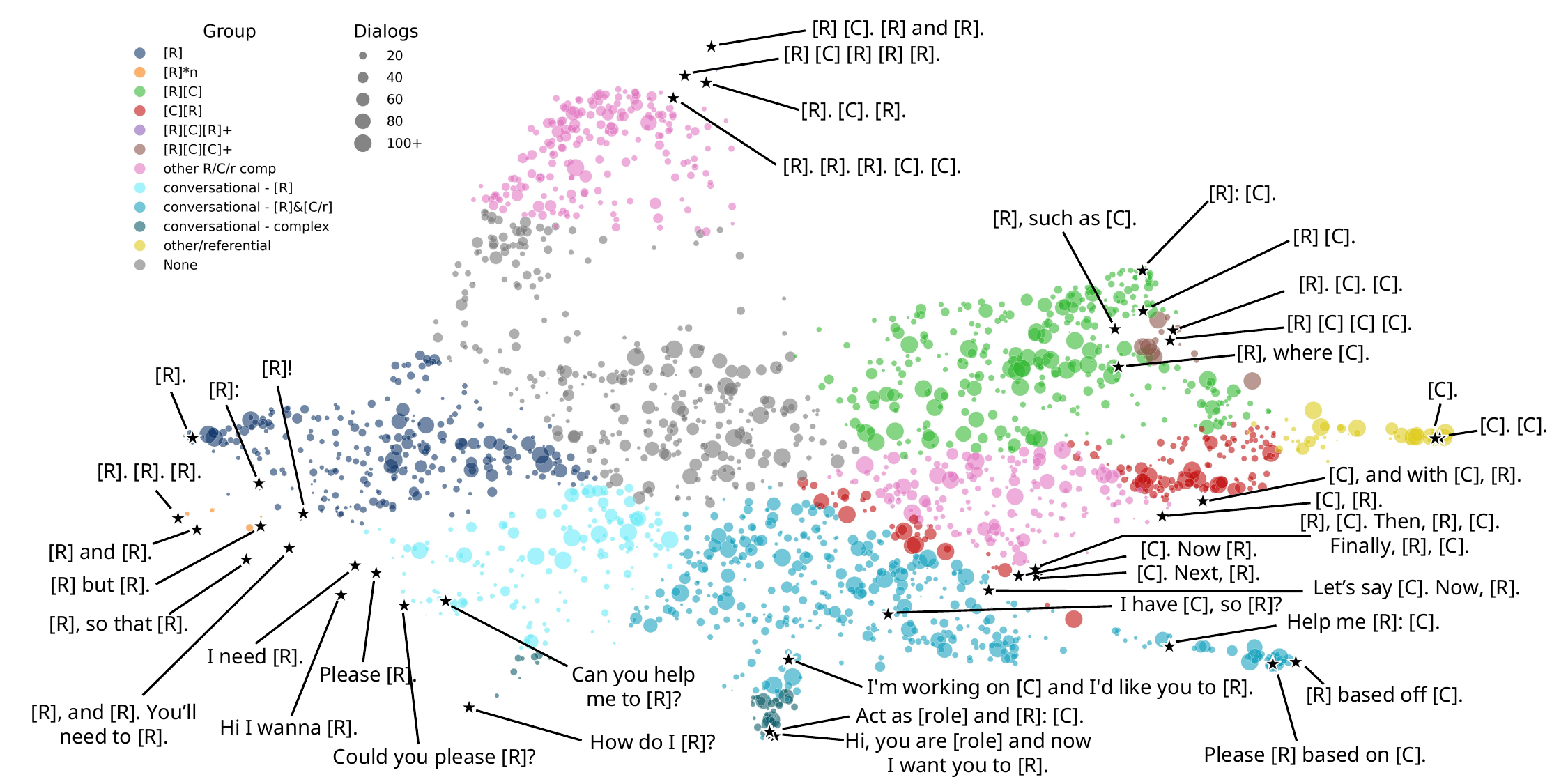}
    \caption{The overview of the ReCCRE dataset as a user-level 2-D plot. Each circle represents a user, with its size matching their total dialogs and color clustered based on the closest anchor point. In general, the horizontal axis displays the ratio of $[R]$ and $[C]$, and the vertical axis ranges from the most to least conversational.}
    \label{fig:PaCMAP_scatter_draft}
\end{figure*}

\begin{figure*}[t]
    \centering
    \includegraphics[width=0.48\linewidth]{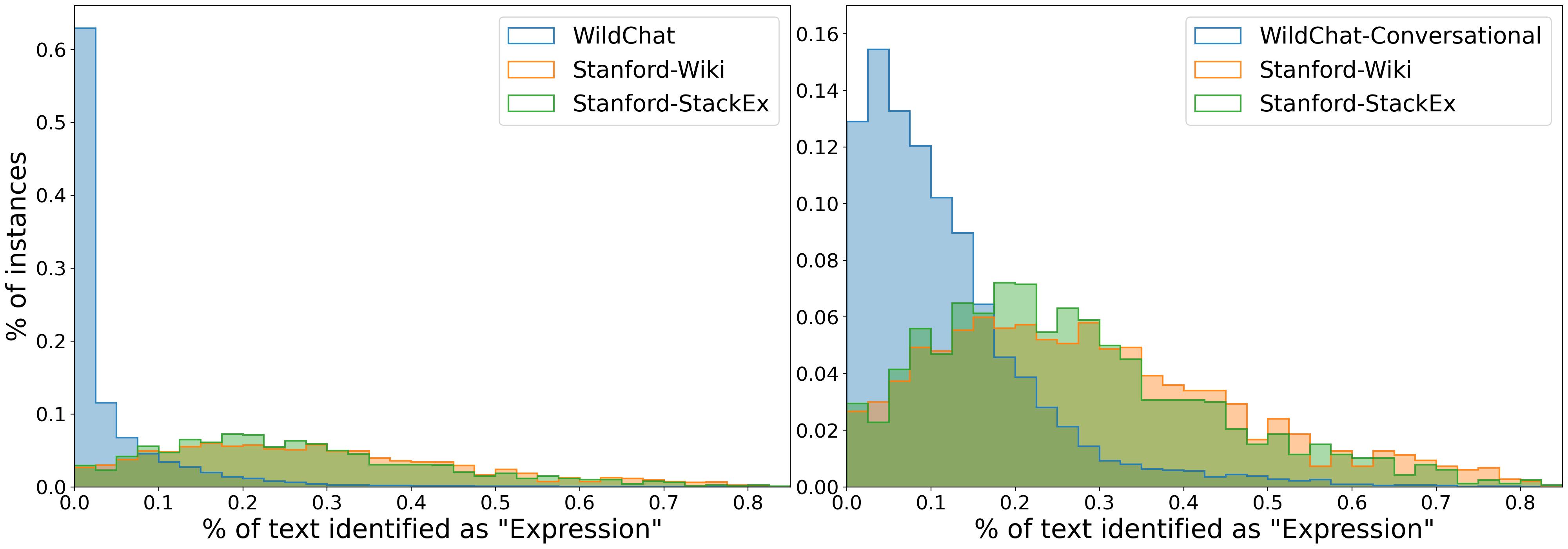} \hfill
    \includegraphics[width=0.48\linewidth]{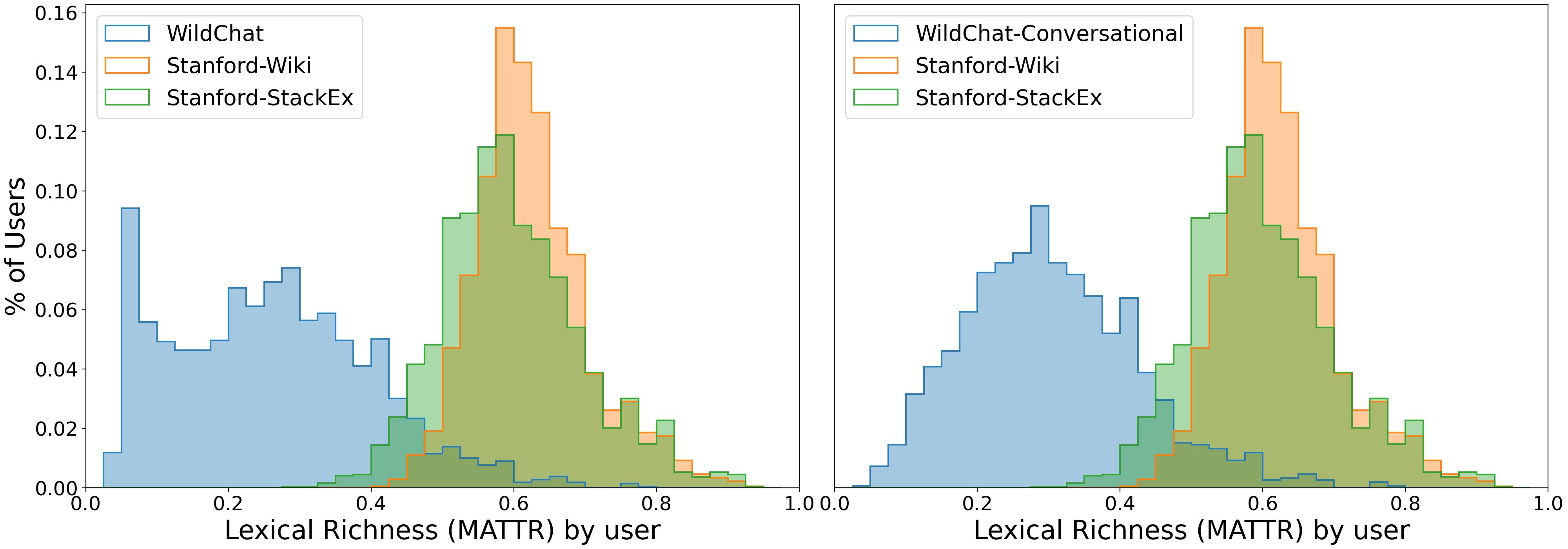}
    \caption{Comparing ReCCRE-WildChat and Stanford Politeness dataset: the amount (\%) of task-independent expressions (left), and the distribution of speaker Lexical Richness (right; measured by Moving-Average Text-Token Ratio). For each metric, the first figure shows the full data and the second the conversational portion only.}
    \label{fig:formatting_percentage_comparison}
\end{figure*}

\subsection{Categorization and Illustration}
\subsubsection{Taxonomy of Expressions}
To understand and generalize the annotations, we start from a taxonomy of expressions in request-making shown in Table~\ref{tab:expression_types}. We refer to an expression as \textit{conversational} if it involves explicit signs of conversing with another party, such as person (``You'' and ``I''), politeness strategies, and greetings. Conversely, if an utterance is a mere imperative combination of $[R]$ and $[C]$ without signs of conversation, it is marked as \textit{non-conversational}. We further break down the expressions based on their structure and complexity, e.g., whether multiple request or context components are involved.

\paragraph{Anchor Points} To calibrate the varied user inputs, we collect 40 different instances of the most frequent expressions in the dataset, covering all 11 categories.
We use them as \textit{anchor points} in our analyses (including subsequent figures) to provide reference for visualizations, and more importantly, to allow categorization of arbitrary unseen expressions based on their closest anchor points.

\subsubsection{Visualizing the data distribution}
We vectorize user expressions with a SotA text-embedding model, \texttt{gte-large-en-v1.5}~\cite{li2023towards}, with $[R]$, $[C]$, and $[role]$ wrapped as formatted placeholder tokens (\texttt{\_\_[REQUEST]\_\_}, etc.)
Each request-making utterance is encoded as a 1024-dim vector, and a user is represented by the average of their utterances.
We then map all users as well as the anchor points to a 2-D space using PaCMAP~\cite{JMLR:v22:20-1061}, a SotA dimension reduction (DR) method.
In this way, the collections of user expressions are depicted as a scatter plot in Figure~\ref{fig:PaCMAP_scatter_draft}, where each bubble represents a long-term user. The bubble size is in proportion to the number of dialogs a user created. Bubbles are colored based on the closest anchor point, or grey if not close enough to any anchor points.

Note that the horizontal layout corresponds to the composition of Goals ($[R]$) and Context ($[C]$), where the leftmost represents the sole $[R]$ and the rightmost corresponds to $[C]$ only, and the combined forms are in between. Meanwhile, the vertical positions can be interpreted as the ``conversationality'': the bottom ones, featuring role assigning and politeness patterns like ``please'', are closest to the expressions and force of natural conversations; the topmost ones, with a repetitive sequence of $[R]$ and $[C]$ with no additional text, is most tool-like and unlikely in human-human request-making.

\subsection{Human-LLM request-making differs from human-human counterparts}
Factoring out case-specific contexts allows us to compare request-making language in LLM chats with human-human interaction.
To evaluate this difference, we apply the same segmentation scheme to the Stanford Politeness Datasets~\cite{danescu-niculescu-mizil-etal-2013-computational}, a prominent pragmatics corpus relevant to request-making.
The corpus records two natural dialog sources, Wikipedia editor discussions (Stanford-Wiki) and the StackExchange Q\&A forum (Stanford-StackEx).
Each utterance by design involves the speaker making requests to another member.
We therefore compare how requests are delivered in the two human-human and the user-LLM scenarios.

We note the distinct composition of the utterances and the Lexical Richness of expressions~\cite{laufer1995vocabulary, lex}.
Figure~\ref{fig:formatting_percentage_comparison} compares (1) the percentage of input text as expression and (2) the user-level Moving-Average Text-Token Ratio~\cite{covington2010cutting} of ReCCRE and Stanford datasets.
The two natural subsets share highly similar stats despite distinct sources and contexts; however, ReCCRE uses qualitatively fewer expressions to embed requests, and user language has much lower diversity.
One major reason is the presence of non-conversational ``imperative'' spans (Table~\ref{tab:expression_types});
e.g., ``Write an article about jogging.'' is marked as a single $[R]$ component with minimal or no formatting.
This is a common paradigm different from natural conversations, as the latter would require appropriate social grounding.
However, if confined to the \textit{conversational} instances (the subplots on the right in both figures) by removing the simple combinations of request content and context, we see that the difference still holds.
In other words, the existence of the non-conversational imperatives is not sufficient to account for the difference; there exist fundamental contrasts of language use specific to the human-LLM scenario, independent from request content.

\begin{figure}[t]
    \centering
    \includegraphics[width=0.95\linewidth]{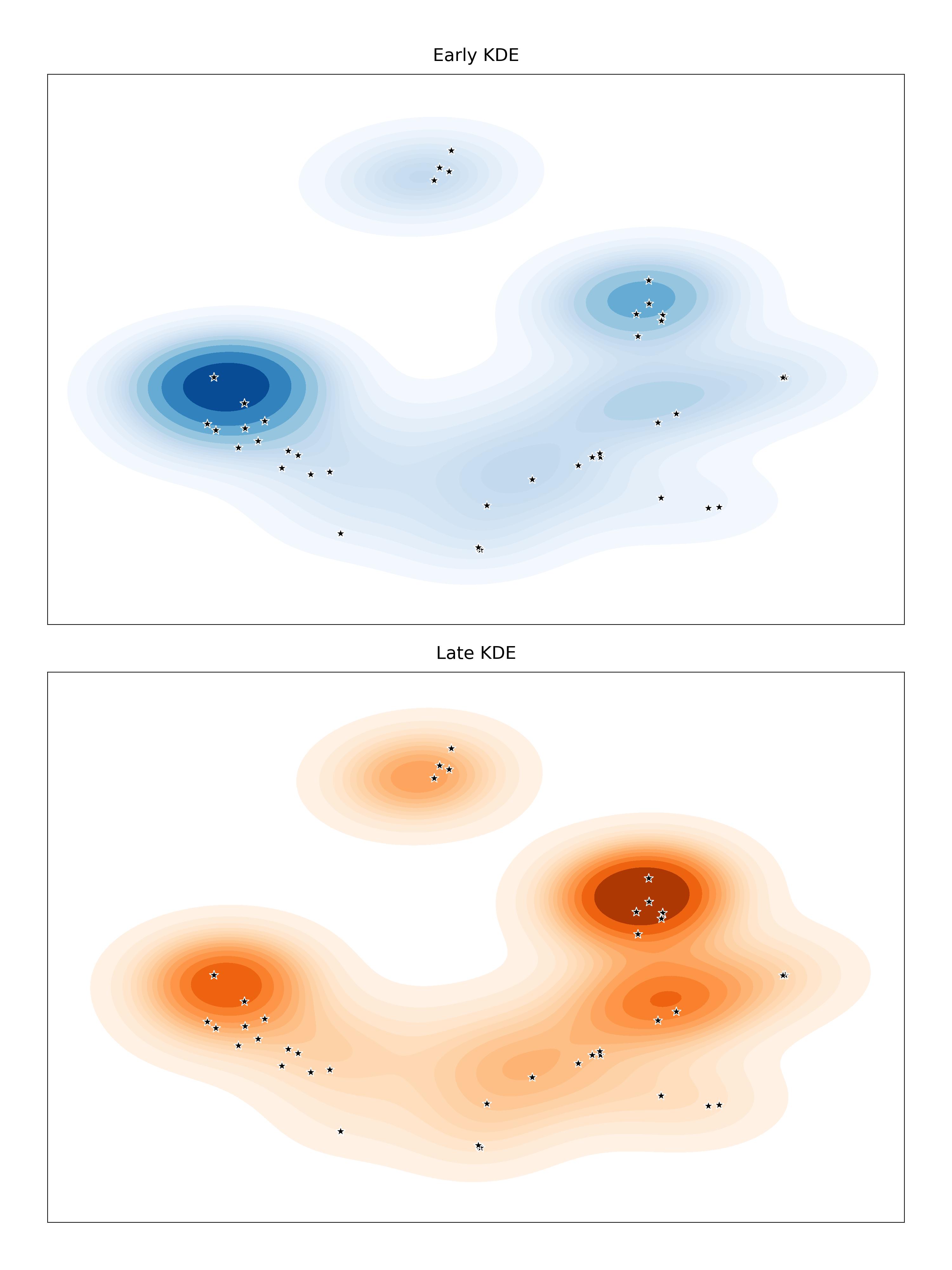}
    \caption{Distribution of the 1st (left) and 20th dialogs (right) of eligible long-term users as Kernel Density Estimation (KDE) plots in the 2D space from Fig.~\ref{fig:PaCMAP_scatter_draft}.}
    \label{fig:early_late_kde}
\end{figure}

\section{Diachronic Analysis of Request-making}
Enabled by the new data infrastructure, we move on to discuss a novel paradigm: modeling user behaviors and patterns in a \textit{diachronic} manner, thereby understanding interaction as a systematic and dynamic process.
We will first inspect the lifecycles of individual users, and discuss how the fundamental properties, such as the diversity of expression types, change across time.
Next, we interpret the holistic evolution patterns formed by the community collectively, and extend further to compare the core user base and the lay public.

\subsection{Modeling User Lifecycle}
Long-term, non-intrusive documentary of user-LLM interactions provides an interface for the user lifecycles~\cite{zhu2025data}, i.e., the full course of actions and use history.
Here, we discuss how a fully text-based analysis can help to model the change of use patterns across time.

\subsubsection{What expressions were used and how are they structured?}
\begin{figure}[t]
    \centering
    \includegraphics[width=0.92\linewidth]{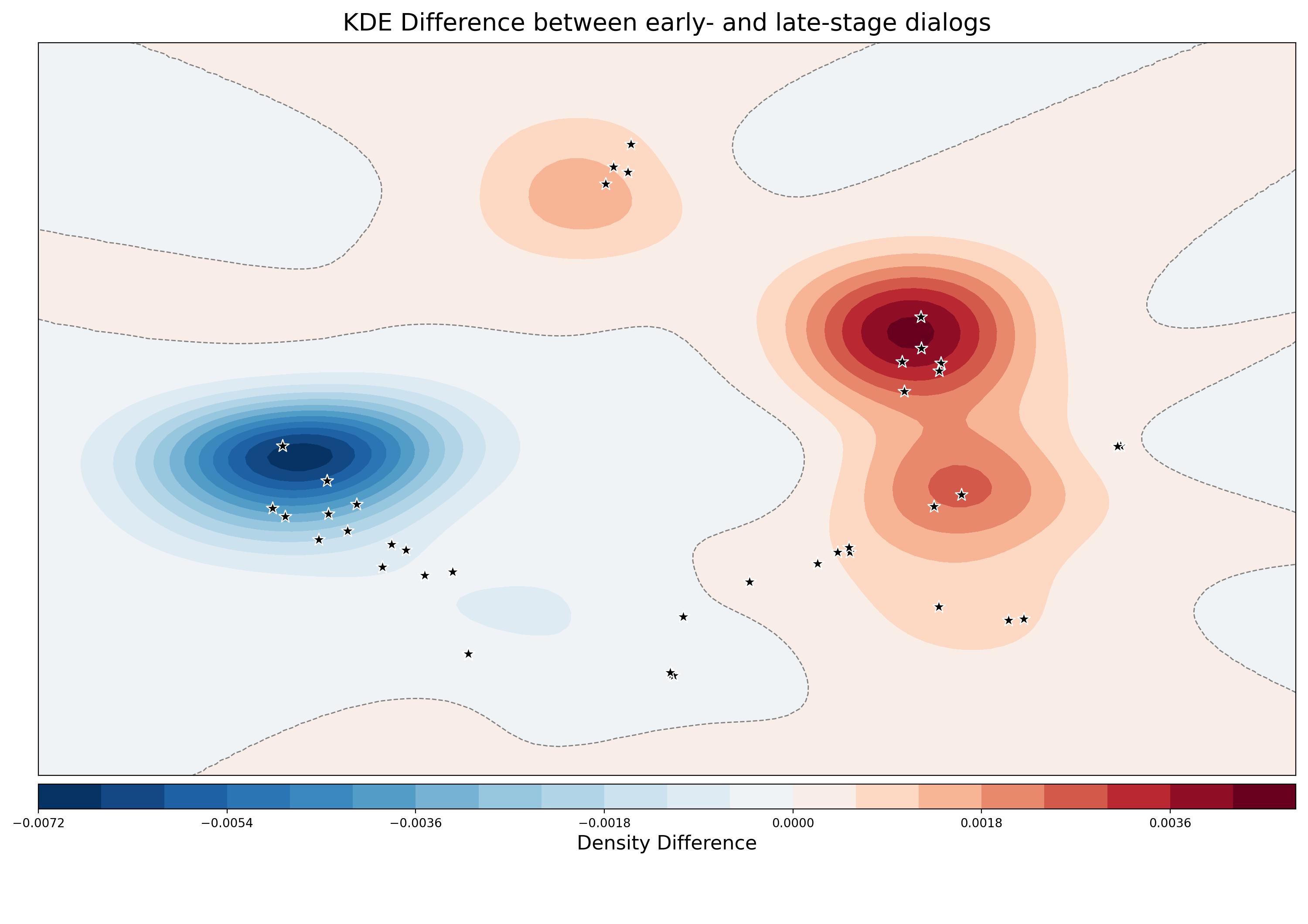}
    \caption{Difference of densities between the early and late inputs of long-term users (Fig.~\ref{fig:early_late_kde}). In the later stage, the balanced combination of $[R]$ and $[C]$ (upper-right, red) sees major inflows, while the non-conversational stacks of $[R]$s (blue, left) significantly decreases.}
    \label{fig:kde_diff}
\end{figure}

As a natural continuation of Figure~\ref{fig:PaCMAP_scatter_draft}, we further add the dimension of time and inspect the most common patterns across individual dialogs at different stages.
To model user lifecycles, we position a pair of early- and late-stage input data from long-term users in the same 2-D space from Fig.~\ref{fig:PaCMAP_scatter_draft} and apply Kernel Density Estimation to model the frequency of user expressions.
Figure~\ref{fig:early_late_kde} shows the comparison of the 1st (left) and the 20th (right) request-making dialog of all eligible long-term users, with the same anchor points from Fig.~\ref{fig:PaCMAP_scatter_draft}, and we directly depict the difference between the two KDEs in Figure~\ref{fig:kde_diff}.
For the users' first explorations, most utterances consist of only one $[R]$ or the simple combination of two. This describes the exploratory stage of interaction with the system with more concise and generic requests.
However, as users gain familiarity, this pattern drastly decreases (though still a major type), and the mass is significantly transitioned to the addition of more specific contexts on the right side, as well as the more complex type of multiple goals and context (top).
This suggests a communal shift towards requests with higher specificity and complexity: Users tend to tailor requests with more context details and fine-grained content later on.
This ``show or tell'' transition reemphasizes that, whereas users may develop distinct use scenarios and tasks, there are indeed fundamental commonalities to be mined at the expression level.

\begin{figure*}[t]
    \centering
    \begin{subfigure}[t]{0.32\textwidth} 
        \centering
        \includegraphics[width=\textwidth]{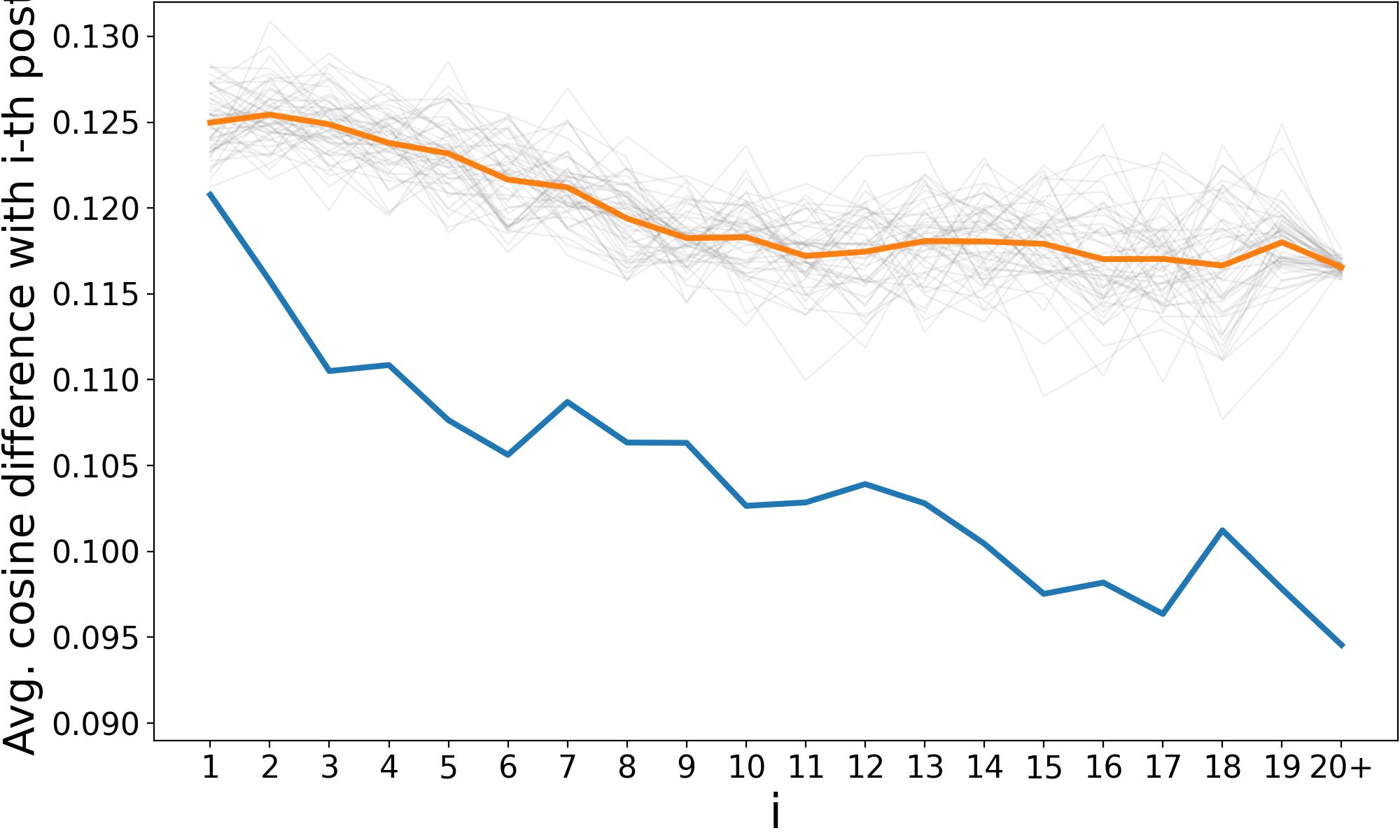}
        \caption{$k=1$}
        \label{fig:LexRich_same-user_window-size-1_comp}
    \end{subfigure} \hfill %
    \begin{subfigure}[t]{0.305\textwidth}
        \centering
        \includegraphics[width=\textwidth]{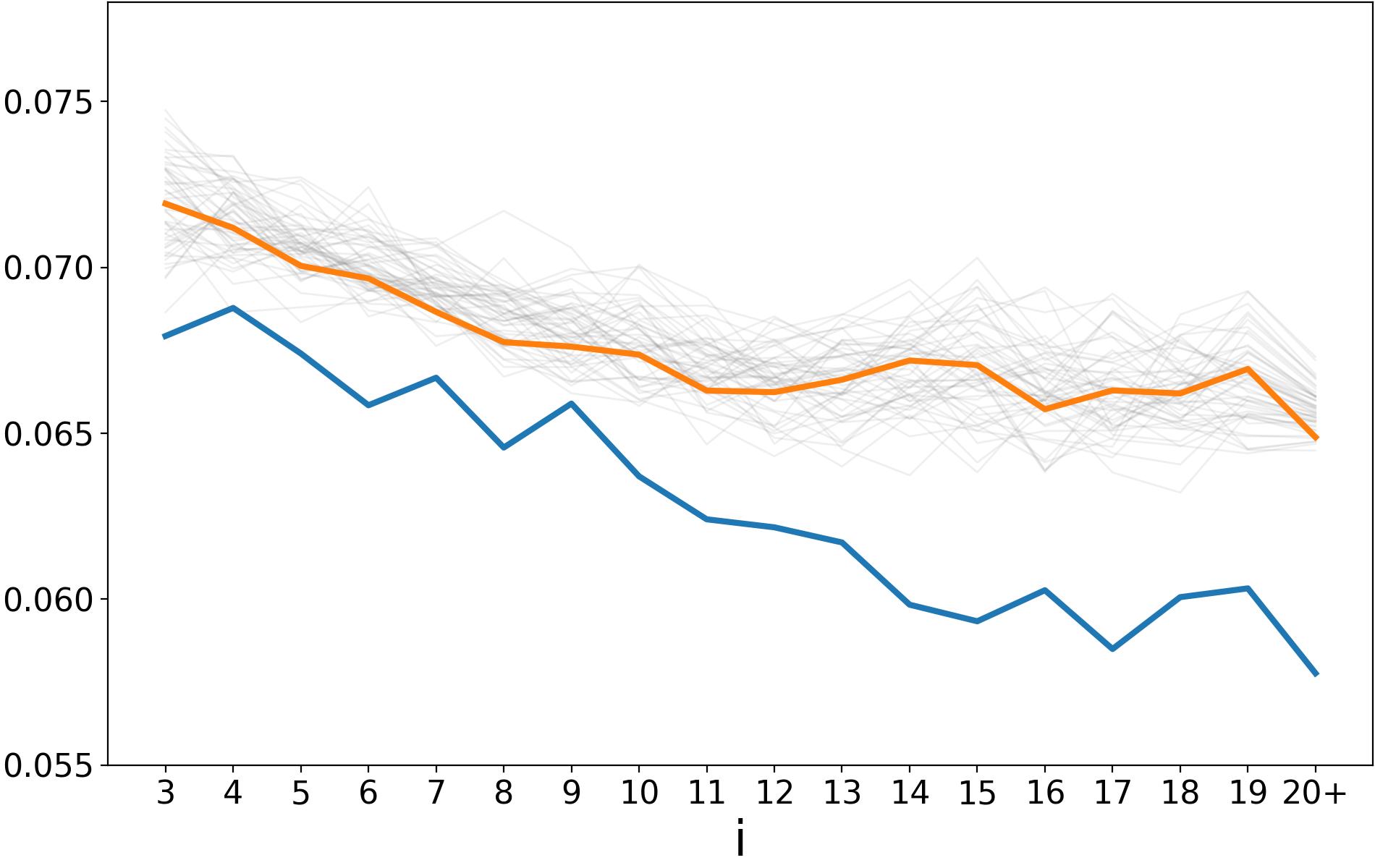}
        \caption{$k=3$}
        \label{fig:LexRich_same-user_window-size-3_comp}
    \end{subfigure} \hfill %
    \begin{subfigure}[t]{0.31\textwidth}
        \centering
        \includegraphics[width=\textwidth]{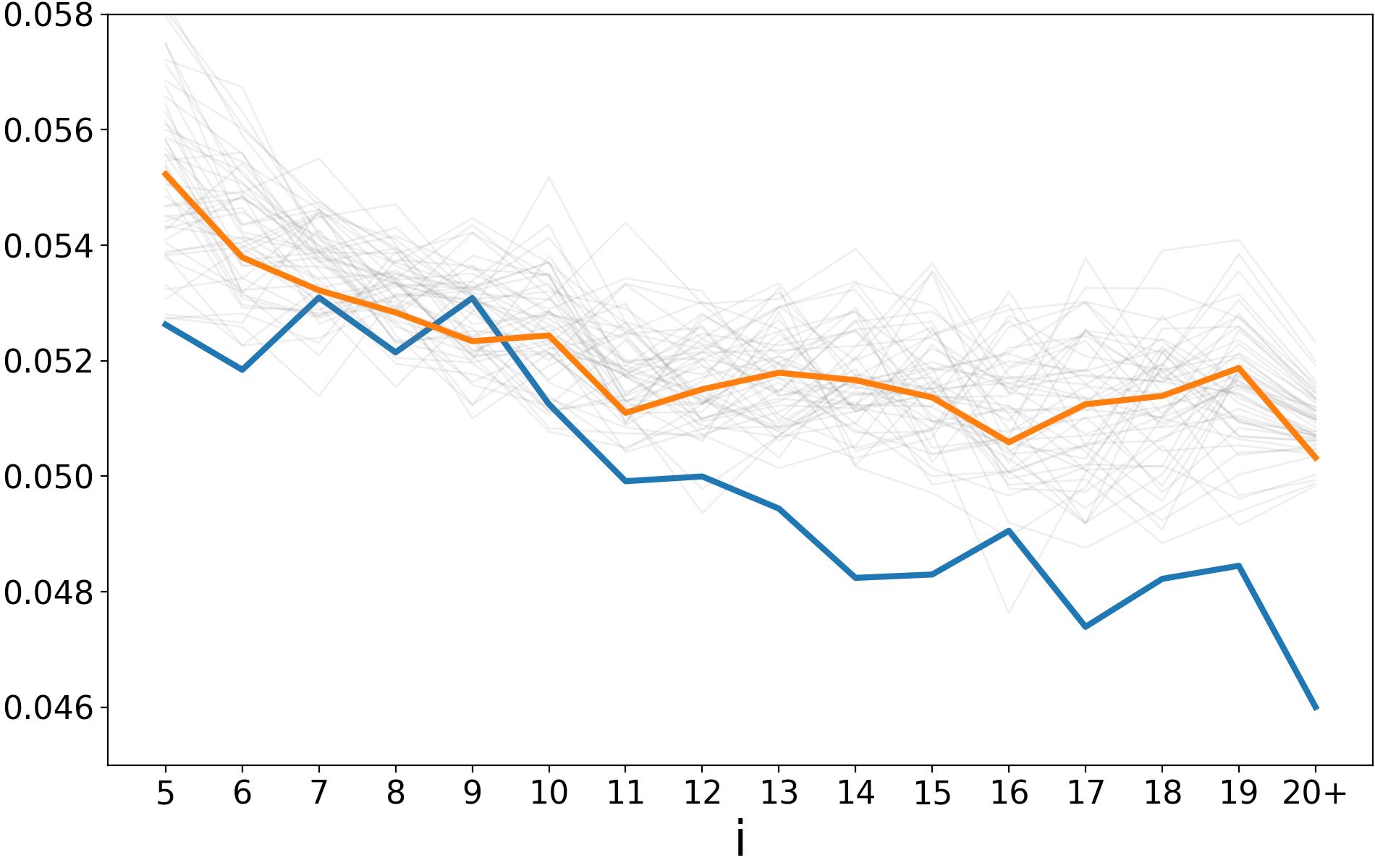}
        \caption{$k=5$}
        \label{fig:LexRich_same-user_window-size-5_comp}
    \end{subfigure}%
    \caption{Convergence of user expressions over time under different window sizes $k$.}
    \label{fig:LexRich_same-user}
\end{figure*}

\subsubsection{User-level Evolutions}
The user expression space like Fig.~\ref{fig:PaCMAP_scatter_draft} enables the study focusing on individual users.
Further, we also wonder if these paths of users can be collected across users to suggest deeper trends.
We consider the effect of familiarity on a user's tendency to repeat the same kinds of expressions they have used previously.
One hypothesis is that a user's expressions may converge with more experience, as they develop their ``go-to'' choices and stick to how that have worked in previous cases.
Alternatively, as users gain familiarity, they may understand the capabilities and boundaries of the LLMs better, and thus rely less on rigorous prompt formats and interact in more casual, arbitrary ways.

To test the contrasting diachronic hypotheses, we examine the minimal difference between a long-term user's input and their most recent inputs, i.e., between their $i$-th and its previous $k$ request utterances,
$\min_{j=1, ..., k-1} [1 - sim(U_{i}, U_{i-j})]$,
for all valid $i$ given window size $k$. Then, we collect the minimal difference across all users and compute the average for each position $i$, to illustrate the averaged step-by-step convergence (or divergence) of expressions within a long-term user's lifecycle.

\begin{figure*}[t]
    \centering
    \includegraphics[width=0.49\textwidth]{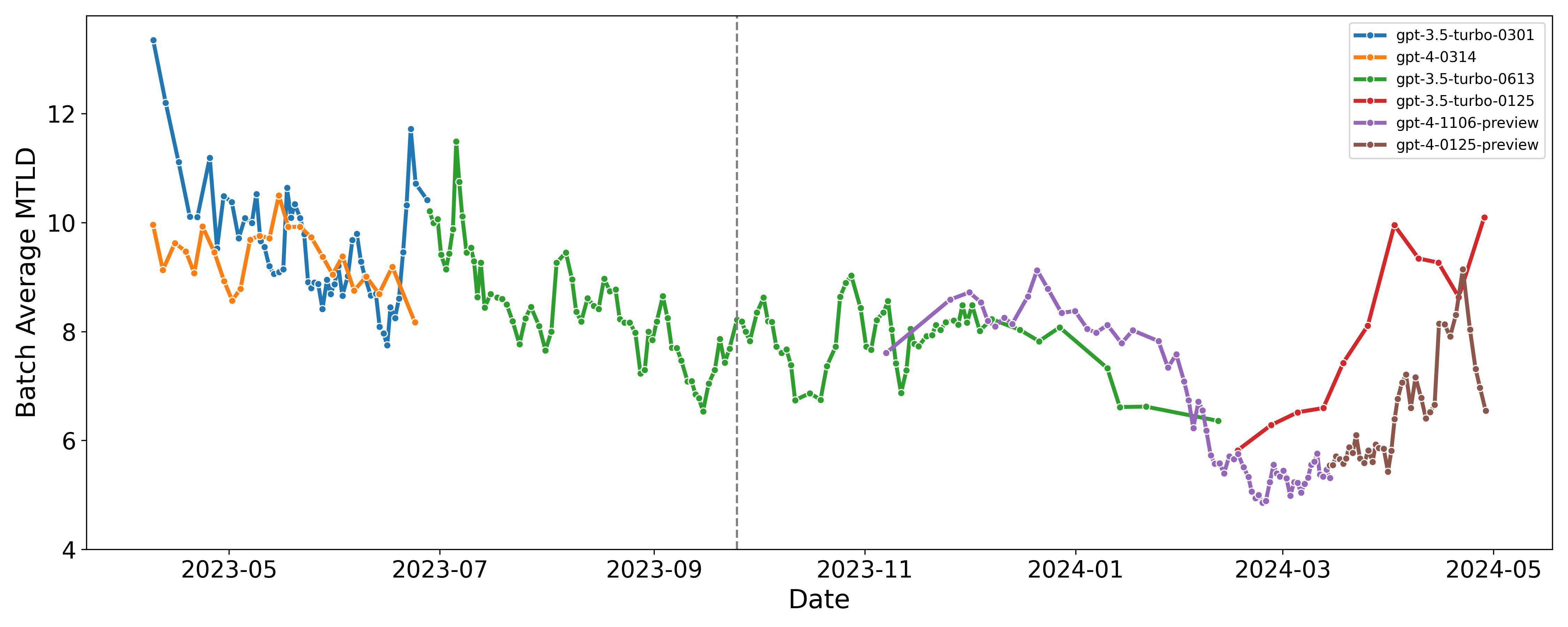} \hfill
    \includegraphics[width=0.49\textwidth]{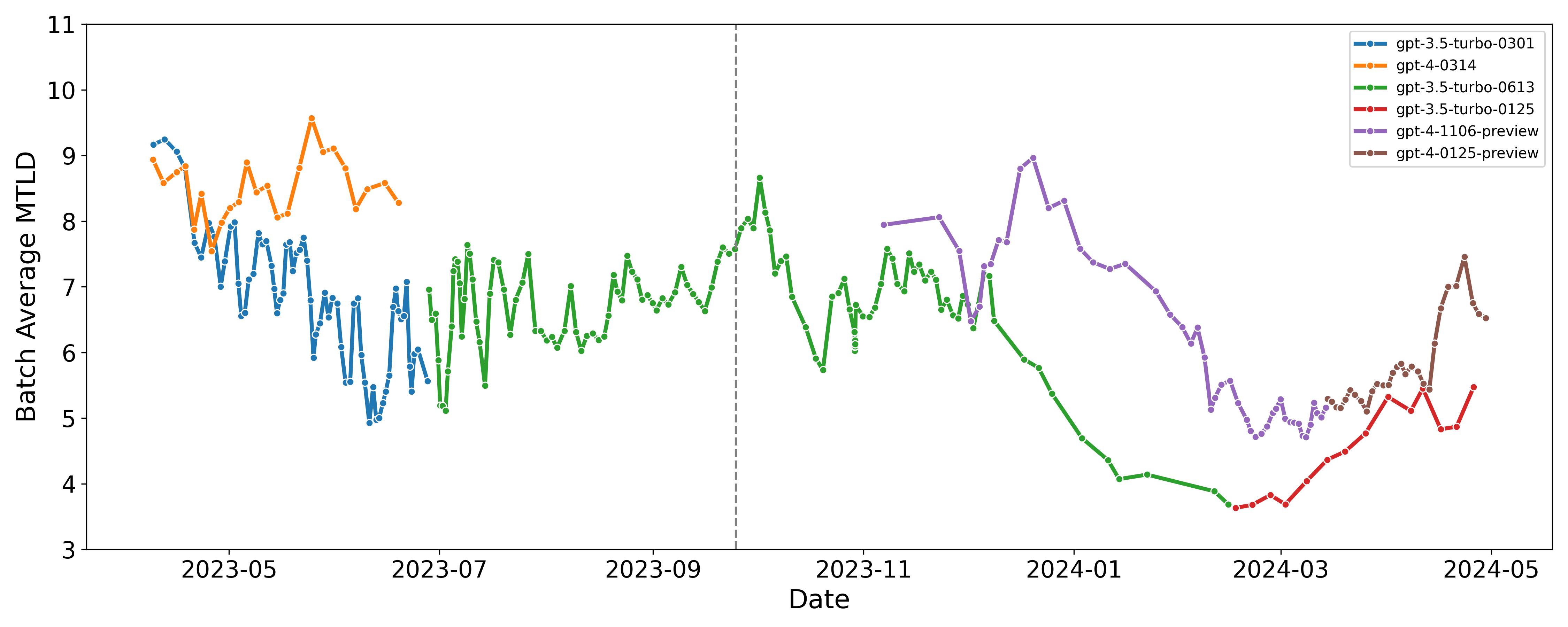}
    \caption{Lexical Richness of long-term users (left) and all users (right) across the full time span.}
    \label{fig:LexRich_time-lapse_all-users}
\end{figure*}

Figure~\ref{fig:LexRich_same-user} displays the results under different window sizes $k$. The actual chronological data is shown in blue, compared against the average of 50 random trials where the same user's dialogs are randomly shuffled (shown in orange), thus breaking diachronic ties.
We observe strong evidence for the \textit{convergence} hypothesis across time: the difference between a new input and its closest predecessors sees a drastic, continuous decline as users continue to interact with the system.
This is quantitatively different from the baseline level of random shuffles, and the difference between the two gets more significant with more input requests. This suggests that users overall develop rather stable patterns of usage as they gain familiarity. 
Meanwhile, we also note that this pattern is most significant with window size $k=1$ (Fig.~\ref{fig:LexRich_same-user_window-size-1_comp}), i.e., when comparing each input with the exact one previous dialog. The gap between real and random situations is reduced with $k=3$ (Fig.~\ref{fig:LexRich_same-user_window-size-3_comp}) and further with $k=5$ (Fig.~\ref{fig:LexRich_same-user_window-size-5_comp}).
This indicates that the most recent cases consistently serve as more significant references for a new user request, whereas the effect of earlier inputs decreases with their lower recency.

\subsection{Exploring and picturing the community}
\label{subsec:community_mtld}

So far we have modeled the evolving request-making expressions from the perspectives of individual users and requests.
We finally discuss a broader horizon of ReCCRE: Is it possible to draw a full landscape of evolving human-LLM interaction from observations, modeling the complete evolution trends in the user community as a whole?

We show that, with metrics as simple as Lexical Richness, we are poised to delineate the climate in great detail and discover the subtleties therein. Specifically, we use MTLD~\cite{mccarthy2005assessment,mccarthy2010mtld} as it's almost fully length-invariant and suitable for random collections.
The intuitive interpretation of batched Lexical Richness is a measurement of expression diversity across all users: A higher richness indicates that there are more distinct presentations of requests, while a lower richness indicates converged, collective ways of interactions and clearer system affordances.

\paragraph{Comparing long-term and lay users} While our analyses have focused on long-term users for the legitimate comparisons over time, the holistic view here enables us to compare the experience of the ``regulars'' and the entire public user base. The MTLD data for the full dataset with all 18,964 users is illustrated in Figure~\ref{fig:LexRich_time-lapse_all-users}, where proportionally larger batch sizes are used to create comparable densities of data points in the two figures.

\subsubsection{Observations}
Both long-term and all users share a similar overall trend. The initial deployment of WildChat saw heavily heterogeneous attempts, especially in users who later became long-term. After initial oscillations, the diversity level of expressions stabilized in the extended period of \texttt{gpt-3.5-turbo-0613}. However, as the most advanced \texttt{gpt-4-1106-preview} was introduced, there seemed a paradigm shift towards a new low. The trend is further different with another major model replacement around March 2024, and both groups see drastically more diverse expressions.

New models, especially the ones from the same family (prefix), take on the exact use patterns of their predecessors. Simultaneous models also see shared evolutions: for instance, the emerging \texttt{gpt-4-1106-preview} leads users to also interact with \texttt{gpt-3.5-turbo-0613} with less diverse expressions, though the latter is no different from before.
Interestingly, long-term users and the lay public seem to show flipped perceptions of model capabilities and usages, as seen in the earliest and latest parts of WildChat. These indicate further complexities yet to be explored, regarding the perception of LLMs in relation to familiarity.

\section{Discussion}

The ReCCRE framework enables us to move beyond surface-level task types to understand the deeper structure of \textit{how} users formulate requests as LLM inquiries.
Analyzing 211k real-world user inputs, we show that human-LLM interaction constitutes a distinct communicative domain with its own evolving linguistic conventions and implications for the broader NLP community.

\paragraph{Understanding the Emergence and Temporal Dynamics of Human-LLM Pragmatics.}
Real-world user inputs are heterogeneous: filler language, role-play, and non-conversational inputs are common.
We also show that users do not transfer existing conversational norms to LLM interactions, but instead develop new registers with distinct pragmatic features. 
Existing work has noted understudied features in user interaction such as under-specification~\cite{mysore2025prototypical}, as well as key contrasts in real-world users such as conversational and contextualized~\cite{malaviya2025contextualized} vs. fragmented or task-oriented~\cite{wang2025talk,sarkar2025conversational}, and perceiving LLMs more as a tool vs. collaborator~\cite{schroeder2024large,gao2024aligning}.
ReCCRE makes these fundamental user patterns visible and enables systematic quantitative analyses, connecting observations with qualitative studies and pragmatics research.
Future work may explore whether certain patterns correlate with more successful outcomes, how these patterns vary across cultures or languages, and how systems can be designed to scaffold communication.

Further, our work is an NLP example of longitudinal user studies with LLMs in the wild~\cite{long2025facilitating,zhu2025data,chamberlain2012research}.
We show users collectively move from request-centric to more context-rich interactions as their familiarity with the system grows. This could suggest new evaluation axes such as convergence rate, and interesting resonances with studies on the actual task content. More broadly, this exemplifies the exploration towards systematic understanding of users' trajectories, revealing how people learn, adapt, and converge, and connecting with rich interdisciplinary work, especially HAI user studies.

\paragraph{Rethinking Alignment via Natural Usage Patterns.}

One of the main goals of post-training and alignment procedures is to adapt language models to respond to user interactions, and they must account for the full spectrum of real-world user expression styles~\cite{don2024naturally,malaviya2025contextualized,gao2024aligning}.
Our decomposition provides a principled way to enhance training data to reflect this diversity by recombining different expression templates with various request types, or modeling the temporal progression from novice to experienced user patterns.
The data supplies \textit{authentic} ingredients for data augmentation under \textit{fine-grained control} and for training reward models that prefer grounded, context-respecting responses.
The ReCCRE components of a query may serve as independent supervision signals: requests signify intent, context supports retrieval-augmented grounding, and expressions show the “non-task” linguistic habits that models must identify and match.
By separating the components of queries, we can create new training recipes and more informative evaluation, such as request-identification accuracy or robustness to noisy expressions.

\paragraph{Conclusions}
Our results show that there are consistent, large-scale patterns in how new and experienced users interact with LLM systems.
These patterns only appear when we separate \textit{how} people ask from \textit{what} they ask for.
While the WildChat dataset provides sufficient proof-of-concept, it also sets a protocol for further study across platforms, modalities, languages, communities, and cultures.
Collectively, these directions position ReCCRE not only as an analytic lens but as a practical toolkit for advancing robust, user-centered NLP.

\section*{Limitations}
While our work seeks to provide new resources and paradigms, the data and analysis work both have practical limitations. Due to the limited capabilities of the 14B LLM and the author as annotator and validator, the ReCCRE dataset is selected from the chat logs with English as the major language in the original dataset. This limits the reliable scope of the work, as cultural and language factors can lead to different interaction patterns. Further, our work is based fully on WildChat, which represents a specific type of data collection practice, a certain group of audience, and a specific time window. While there are no additional confounders introduced, it may still inherit biases present in the underlying WildChat logs, such as demographic skews or designs of its UI layout (HugginFace). It is also possible that findings are different in interesting ways in other documented chat logs, such as LMSys~\cite{zhenglmsys,chiang2024chatbot} collected earlier with a different protocol. In our analyses, we largely utilize off-the-shelf metrics including the \texttt{gte} embeddings and Lexical Richness measurements. This is by design, as we would like to demonstrate the usability and low barrier of the dataset. However, we also note that this might limit the boundaries and depth of data analyses, and we encourage further explorations with the ReCCRE data, e.g., fine-tuning models with the resource.

\bibliography{custom}

\begin{thebibliography}{42}
\providecommand{\natexlab}[1]{#1}

\bibitem[{Chamberlain et~al.(2012)Chamberlain, Crabtree, Rodden, Jones, and Rogers}]{chamberlain2012research}
Alan Chamberlain, Andy Crabtree, Tom Rodden, Matt Jones, and Yvonne Rogers. 2012.
\newblock Research in the wild: understanding'in the wild'approaches to design and development.
\newblock In \emph{Proceedings of the designing interactive systems conference}, pages 795--796.

\bibitem[{Cheng et~al.(2025)Cheng, Ghate, Hua, Wang, Shen, and Fang}]{cheng2025realm}
Jingwen Cheng, Kshitish Ghate, Wenyue Hua, William~Yang Wang, Hong Shen, and Fei Fang. 2025.
\newblock Realm: A dataset of real-world llm use cases.
\newblock \emph{arXiv preprint arXiv:2503.18792}.

\bibitem[{Chiang et~al.(2024)Chiang, Zheng, Sheng, Angelopoulos, Li, Li, Zhu, Zhang, Jordan, Gonzalez et~al.}]{chiang2024chatbot}
Wei-Lin Chiang, Lianmin Zheng, Ying Sheng, Anastasios~Nikolas Angelopoulos, Tianle Li, Dacheng Li, Banghua Zhu, Hao Zhang, Michael Jordan, Joseph~E Gonzalez, and 1 others. 2024.
\newblock Chatbot arena: An open platform for evaluating llms by human preference.
\newblock In \emph{Forty-first International Conference on Machine Learning}.

\bibitem[{Choi et~al.(2024)Choi, Akter, Singh, and Anastasopoulos}]{choi-etal-2024-llm}
Alexander~S. Choi, Syeda~Sabrina Akter, JP~Singh, and Antonios Anastasopoulos. 2024.
\newblock \href {https://doi.org/10.18653/v1/2024.emnlp-main.1230} {The {LLM} effect: Are humans truly using {LLM}s, or are they being influenced by them instead?}
\newblock In \emph{Proceedings of the 2024 Conference on Empirical Methods in Natural Language Processing}, pages 22032--22054, Miami, Florida, USA. Association for Computational Linguistics.

\bibitem[{Covington and McFall(2010)}]{covington2010cutting}
Michael~A Covington and Joe~D McFall. 2010.
\newblock Cutting the gordian knot: The moving-average type--token ratio (mattr).
\newblock \emph{Journal of quantitative linguistics}, 17(2):94--100.

\bibitem[{Danescu-Niculescu-Mizil et~al.(2013)Danescu-Niculescu-Mizil, Sudhof, Jurafsky, Leskovec, and Potts}]{danescu-niculescu-mizil-etal-2013-computational}
Cristian Danescu-Niculescu-Mizil, Moritz Sudhof, Dan Jurafsky, Jure Leskovec, and Christopher Potts. 2013.
\newblock \href {https://aclanthology.org/P13-1025/} {A computational approach to politeness with application to social factors}.
\newblock In \emph{Proceedings of the 51st Annual Meeting of the Association for Computational Linguistics (Volume 1: Long Papers)}, pages 250--259, Sofia, Bulgaria. Association for Computational Linguistics.

\bibitem[{Don-Yehiya et~al.(2024)Don-Yehiya, Choshen, and Abend}]{don2024naturally}
Shachar Don-Yehiya, Leshem Choshen, and Omri Abend. 2024.
\newblock Naturally occurring feedback is common, extractable and useful.
\newblock \emph{arXiv preprint arXiv:2407.10944}.

\bibitem[{Gao et~al.(2024{\natexlab{a}})Gao, Taymanov, Salinas, Mineiro, and Misra}]{gao2024aligning}
Ge~Gao, Alexey Taymanov, Eduardo Salinas, Paul Mineiro, and Dipendra Misra. 2024{\natexlab{a}}.
\newblock \href {https://openreview.net/forum?id=DlYNGpCuwa} {Aligning {LLM} agents by learning latent preference from user edits}.
\newblock In \emph{The Thirty-eighth Annual Conference on Neural Information Processing Systems}.

\bibitem[{Gao et~al.(2024{\natexlab{b}})Gao, Gebreegziabher, Choo, Li, Perrault, and Malone}]{10.1145/3613905.3650786}
Jie Gao, Simret~Araya Gebreegziabher, Kenny Tsu~Wei Choo, Toby Jia-Jun Li, Simon~Tangi Perrault, and Thomas~W Malone. 2024{\natexlab{b}}.
\newblock \href {https://doi.org/10.1145/3613905.3650786} {A taxonomy for human-llm interaction modes: An initial exploration}.
\newblock In \emph{Extended Abstracts of the CHI Conference on Human Factors in Computing Systems}, CHI EA '24, New York, NY, USA. Association for Computing Machinery.

\bibitem[{Handa et~al.(2025)Handa, Tamkin, McCain, Huang, Durmus, Heck, Mueller, Hong, Ritchie, Belonax et~al.}]{handa2025economic}
Kunal Handa, Alex Tamkin, Miles McCain, Saffron Huang, Esin Durmus, Sarah Heck, Jared Mueller, Jerry Hong, Stuart Ritchie, Tim Belonax, and 1 others. 2025.
\newblock Which economic tasks are performed with ai? evidence from millions of claude conversations.
\newblock \emph{arXiv preprint arXiv:2503.04761}.

\bibitem[{He et~al.(2025)He, Naphade, and Huang}]{10.1145/3706598.3714319}
Zeyu He, Saniya Naphade, and Ting-Hao~Kenneth Huang. 2025.
\newblock \href {https://doi.org/10.1145/3706598.3714319} {Prompting in the dark: Assessing human performance in prompt engineering for data labeling when gold labels are absent}.
\newblock In \emph{Proceedings of the 2025 CHI Conference on Human Factors in Computing Systems}, CHI '25, New York, NY, USA. Association for Computing Machinery.

\bibitem[{Huang et~al.(2024)Huang, Van~Rijn, Sucholutsky, Marjieh, and Jacoby}]{huang-etal-2024-characterizing}
Dun-Ming Huang, Pol Van~Rijn, Ilia Sucholutsky, Raja Marjieh, and Nori Jacoby. 2024.
\newblock \href {https://doi.org/10.18653/v1/2024.acl-long.565} {Characterizing similarities and divergences in conversational tones in humans and {LLM}s by sampling with people}.
\newblock In \emph{Proceedings of the 62nd Annual Meeting of the Association for Computational Linguistics (Volume 1: Long Papers)}, pages 10486--10512, Bangkok, Thailand. Association for Computational Linguistics.

\bibitem[{Huang et~al.(2025)Huang, Durmus, McCain, Handa, Tamkin, Hong, Stern, Somani, Zhang, and Ganguli}]{huang2025values}
Saffron Huang, Esin Durmus, Miles McCain, Kunal Handa, Alex Tamkin, Jerry Hong, Michael Stern, Arushi Somani, Xiuruo Zhang, and Deep Ganguli. 2025.
\newblock Values in the wild: Discovering and analyzing values in real-world language model interactions.
\newblock \emph{arXiv preprint arXiv:2504.15236}.

\bibitem[{Ivey et~al.(2024)Ivey, Kumar, Liu, Shen, Rakshit, Raju, Zhang, Ananthasubramaniam, Kim, Yi et~al.}]{ivey2024real}
Jonathan Ivey, Shivani Kumar, Jiayu Liu, Hua Shen, Sushrita Rakshit, Rohan Raju, Haotian Zhang, Aparna Ananthasubramaniam, Junghwan Kim, Bowen Yi, and 1 others. 2024.
\newblock Real or robotic? assessing whether llms accurately simulate qualities of human responses in dialogue.
\newblock \emph{arXiv preprint arXiv:2409.08330}.

\bibitem[{Jin et~al.(2025)Jin, Liu, Li, Zhao, and Qu}]{jin2025jailbreakhunter}
Zhihua Jin, Shiyi Liu, Haotian Li, Xun Zhao, and Huamin Qu. 2025.
\newblock Jailbreakhunter: a visual analytics approach for jailbreak prompts discovery from large-scale human-llm conversational datasets.
\newblock \emph{IEEE Transactions on Visualization and Computer Graphics}.

\bibitem[{Kirk et~al.(2024)Kirk, Whitefield, Rottger, Bean, Margatina, Mosquera-Gomez, Ciro, Bartolo, Williams, He et~al.}]{kirk2024prism}
Hannah~Rose Kirk, Alexander Whitefield, Paul Rottger, Andrew~M Bean, Katerina Margatina, Rafael Mosquera-Gomez, Juan Ciro, Max Bartolo, Adina Williams, He~He, and 1 others. 2024.
\newblock The prism alignment dataset: What participatory, representative and individualised human feedback reveals about the subjective and multicultural alignment of large language models.
\newblock \emph{Advances in Neural Information Processing Systems}, 37:105236--105344.

\bibitem[{Kolawole et~al.(2025)Kolawole, Santhanam, Smith, and Thaker}]{kolawole2025parallelprompt}
Steven Kolawole, Keshav Santhanam, Virginia Smith, and Pratiksha Thaker. 2025.
\newblock Parallelprompt: Extracting parallelism from large language model queries.
\newblock \emph{arXiv preprint arXiv:2506.18728}.

\bibitem[{Laufer and Nation(1995)}]{laufer1995vocabulary}
Batia Laufer and Paul Nation. 1995.
\newblock Vocabulary size and use: Lexical richness in l2 written production.
\newblock \emph{Applied linguistics}, 16(3):307--322.

\bibitem[{Lee et~al.(2025)Lee, Maharana, Pujara, Ren, and Barbieri}]{lee2025realtalk}
Dong-Ho Lee, Adyasha Maharana, Jay Pujara, Xiang Ren, and Francesco Barbieri. 2025.
\newblock Realtalk: A 21-day real-world dataset for long-term conversation.
\newblock \emph{arXiv preprint arXiv:2502.13270}.

\bibitem[{Li et~al.(2023)Li, Zhang, Zhang, Long, Xie, and Zhang}]{li2023towards}
Zehan Li, Xin Zhang, Yanzhao Zhang, Dingkun Long, Pengjun Xie, and Meishan Zhang. 2023.
\newblock Towards general text embeddings with multi-stage contrastive learning.
\newblock \emph{arXiv preprint arXiv:2308.03281}.

\bibitem[{Long et~al.(2025)Long, Wang, Fabre, Wang, Sathya, Wu, Petridis, Li, Chakrabarty, Jiang et~al.}]{long2025facilitating}
Tao Long, Sitong Wang, {\'E}milie Fabre, Tony Wang, Anup Sathya, Jason Wu, Savvas~Dimitrios Petridis, Ding Li, Tuhin Chakrabarty, Yue Jiang, and 1 others. 2025.
\newblock Facilitating longitudinal interaction studies of ai systems.
\newblock In \emph{Adjunct Proceedings of the 38th Annual ACM Symposium on User Interface Software and Technology}, pages 1--5.

\bibitem[{Ma et~al.(2024)Ma, Mei, Gajos, and Arawjo}]{10.1145/3613905.3651100}
Zilin Ma, Yiyang Mei, Krzysztof~Z. Gajos, and Ian Arawjo. 2024.
\newblock \href {https://doi.org/10.1145/3613905.3651100} {Schr\"{o}dinger's update: User perceptions of uncertainties in proprietary large language model updates}.
\newblock In \emph{Extended Abstracts of the CHI Conference on Human Factors in Computing Systems}, CHI EA '24, New York, NY, USA. Association for Computing Machinery.

\bibitem[{Malaviya et~al.(2025)Malaviya, Chang, Roth, Iyyer, Yatskar, and Lo}]{malaviya2025contextualized}
Chaitanya Malaviya, Joseph~Chee Chang, Dan Roth, Mohit Iyyer, Mark Yatskar, and Kyle Lo. 2025.
\newblock Contextualized evaluations: Judging language model responses to underspecified queries.
\newblock \emph{Transactions of the Association for Computational Linguistics}, 13:878--900.

\bibitem[{McCarthy(2005)}]{mccarthy2005assessment}
Philip~M McCarthy. 2005.
\newblock \emph{An assessment of the range and usefulness of lexical diversity measures and the potential of the measure of textual, lexical diversity (MTLD)}.
\newblock Ph.D. thesis, The University of Memphis.

\bibitem[{McCarthy and Jarvis(2010)}]{mccarthy2010mtld}
Philip~M McCarthy and Scott Jarvis. 2010.
\newblock Mtld, vocd-d, and hd-d: A validation study of sophisticated approaches to lexical diversity assessment.
\newblock \emph{Behavior research methods}, 42(2):381--392.

\bibitem[{Mireshghallah et~al.(2024)Mireshghallah, Antoniak, More, Choi, and Farnadi}]{mireshghallahtrust}
Niloofar Mireshghallah, Maria Antoniak, Yash More, Yejin Choi, and Golnoosh Farnadi. 2024.
\newblock Trust no bot: Discovering personal disclosures in human-llm conversations in the wild.
\newblock In \emph{First Conference on Language Modeling}.

\bibitem[{Mishra et~al.(2022)Mishra, Firdaus, and Ekbal}]{mishra2022please}
Kshitij Mishra, Mauajama Firdaus, and Asif Ekbal. 2022.
\newblock Please be polite: Towards building a politeness adaptive dialogue system for goal-oriented conversations.
\newblock \emph{Neurocomputing}, 494:242--254.

\bibitem[{Mott et~al.(2024)Mott, Fanganello, and Williams}]{mott2024thing}
Terran Mott, Aaron Fanganello, and Tom Williams. 2024.
\newblock What a thing to say! which linguistic politeness strategies should robots use in noncompliance interactions?
\newblock In \emph{Proceedings of the 2024 ACM/IEEE International Conference on Human-Robot Interaction}, pages 501--510.

\bibitem[{Mysore et~al.(2025)Mysore, Das, Cao, and Sarrafzadeh}]{mysore2025prototypical}
Sheshera Mysore, Debarati Das, Hancheng Cao, and Bahareh Sarrafzadeh. 2025.
\newblock Prototypical human-ai collaboration behaviors from llm-assisted writing in the wild.
\newblock \emph{arXiv preprint arXiv:2505.16023}.

\bibitem[{Sarkar et~al.(2025)Sarkar, Sarrafzadeh, Chandrasekaran, Rangan, Resnik, Yang, and Jauhar}]{sarkar2025conversational}
Rupak Sarkar, Bahareh Sarrafzadeh, Nirupama Chandrasekaran, Nagu Rangan, Philip Resnik, Longqi Yang, and Sujay~Kumar Jauhar. 2025.
\newblock Conversational user-ai intervention: A study on prompt rewriting for improved llm response generation.
\newblock \emph{arXiv preprint arXiv:2503.16789}.

\bibitem[{Schroeder et~al.(2024)Schroeder, Aubin Le~Qu{\'e}r{\'e}, Randazzo, Mimno, and Schoenebeck}]{schroeder2024large}
Hope Schroeder, Marianne Aubin Le~Qu{\'e}r{\'e}, Casey Randazzo, David Mimno, and Sarita Schoenebeck. 2024.
\newblock Large language models in qualitative research: Can we do the data justice?
\newblock \emph{arXiv e-prints}, pages arXiv--2410.

\bibitem[{Shen(2022)}]{lex}
Lucas Shen. 2022.
\newblock \href {https://doi.org/10.5281/zenodo.6607007} {{LexicalRichness: A small module to compute textual lexical richness}}.

\bibitem[{Tamkin et~al.(2024)Tamkin, McCain, Handa, Durmus, Lovitt, Rathi, Huang, Mountfield, Hong, Ritchie et~al.}]{tamkin2024clio}
Alex Tamkin, Miles McCain, Kunal Handa, Esin Durmus, Liane Lovitt, Ankur Rathi, Saffron Huang, Alfred Mountfield, Jerry Hong, Stuart Ritchie, and 1 others. 2024.
\newblock Clio: Privacy-preserving insights into real-world ai use.
\newblock \emph{arXiv preprint arXiv:2412.13678}.

\bibitem[{Wang et~al.(2025)Wang, Wang, and Qiu}]{wang2025talk}
Xiaoyi Wang, Yuran Wang, and Xingyi Qiu. 2025.
\newblock How to talk to ai: The role of preset prompt language styles in shaping conversational experience.
\newblock \emph{International Journal of Human--Computer Interaction}, 41(12):7763--7778.

\bibitem[{Wang et~al.(2021)Wang, Huang, Rudin, and Shaposhnik}]{JMLR:v22:20-1061}
Yingfan Wang, Haiyang Huang, Cynthia Rudin, and Yaron Shaposhnik. 2021.
\newblock \href {http://jmlr.org/papers/v22/20-1061.html} {Understanding how dimension reduction tools work: An empirical approach to deciphering t-sne, umap, trimap, and pacmap for data visualization}.
\newblock \emph{Journal of Machine Learning Research}, 22(201):1--73.

\bibitem[{Zamfirescu-Pereira et~al.(2023)Zamfirescu-Pereira, Wong, Hartmann, and Yang}]{10.1145/3544548.3581388}
J.D. Zamfirescu-Pereira, Richmond~Y. Wong, Bjoern Hartmann, and Qian Yang. 2023.
\newblock \href {https://doi.org/10.1145/3544548.3581388} {Why johnny can’t prompt: How non-ai experts try (and fail) to design llm prompts}.
\newblock In \emph{Proceedings of the 2023 CHI Conference on Human Factors in Computing Systems}, CHI '23, New York, NY, USA. Association for Computing Machinery.

\bibitem[{Zhang et~al.(2025{\natexlab{a}})Zhang, Zhu, Yang, Tseng, Jiang, and Rzeszotarski}]{zhang2025navigating}
Chao Zhang, Shengqi Zhu, Xinyu Yang, Yu-Chia Tseng, Shenrong Jiang, and Jeffrey~M Rzeszotarski. 2025{\natexlab{a}}.
\newblock Navigating the fog: How university students recalibrate sensemaking practices to address plausible falsehoods in llm outputs.
\newblock In \emph{Proceedings of the 7th ACM Conference on Conversational User Interfaces}, pages 1--15.

\bibitem[{Zhang et~al.(2025{\natexlab{b}})Zhang, Liew, and Piumsomboon}]{zhang2025one}
Zhouqing Zhang, Kongmeng Liew, and Tham Piumsomboon. 2025{\natexlab{b}}.
\newblock What one million prompts tells us about ai usage, topics, and preferences.
\newblock In \emph{2025 IEEE Conference on Artificial Intelligence (CAI)}, pages 174--179. IEEE.

\bibitem[{Zhao et~al.(2024)Zhao, Ren, Hessel, Cardie, Choi, and Deng}]{zhaowildchat}
Wenting Zhao, Xiang Ren, Jack Hessel, Claire Cardie, Yejin Choi, and Yuntian Deng. 2024.
\newblock Wildchat: 1m chatgpt interaction logs in the wild.
\newblock In \emph{The Twelfth International Conference on Learning Representations}.

\bibitem[{Zheng et~al.(2024{\natexlab{a}})Zheng, Chiang, Sheng, Li, Zhuang, Wu, Zhuang, Li, Lin, Xing et~al.}]{zhenglmsys}
Lianmin Zheng, Wei-Lin Chiang, Ying Sheng, Tianle Li, Siyuan Zhuang, Zhanghao Wu, Yonghao Zhuang, Zhuohan Li, Zi~Lin, Eric Xing, and 1 others. 2024{\natexlab{a}}.
\newblock Lmsys-chat-1m: A large-scale real-world llm conversation dataset.
\newblock In \emph{The Twelfth International Conference on Learning Representations}.

\bibitem[{Zheng et~al.(2024{\natexlab{b}})Zheng, Pei, Logeswaran, Lee, and Jurgens}]{zheng-etal-2024-helpful}
Mingqian Zheng, Jiaxin Pei, Lajanugen Logeswaran, Moontae Lee, and David Jurgens. 2024{\natexlab{b}}.
\newblock \href {https://doi.org/10.18653/v1/2024.findings-emnlp.888} {When ``a helpful assistant'' is not really helpful: Personas in system prompts do not improve performances of large language models}.
\newblock In \emph{Findings of the Association for Computational Linguistics: EMNLP 2024}, pages 15126--15154, Miami, Florida, USA. Association for Computational Linguistics.

\bibitem[{Zhu et~al.(2025)Zhu, Rzeszotarski, and Mimno}]{zhu2025data}
Shengqi Zhu, Jeffrey~M Rzeszotarski, and David Mimno. 2025.
\newblock Data paradigms in the era of llms: On the opportunities and challenges of qualitative data in the wild.
\newblock In \emph{Proceedings of the Extended Abstracts of the CHI Conference on Human Factors in Computing Systems}, pages 1--8.

\end{thebibliography}

\appendix

\end{document}